# Convolutional Dictionary Pair Learning Network for Image Representation Learning

**Zhao Zhang**[1,2] and **Yulin Sun**[1] and **Yang Wang**[2] and **Zhengjun Zha**[3] and **Shuicheng Yan**[4] and **Meng Wang**[2]

**Abstract.** Both the Dictionary Learning (DL) and Convolutional Neural Networks (CNN) are powerful image representation learning systems based on different mechanisms and principles, however whether we can seamlessly integrate them to improve the performance is noteworthy exploring. To address this issue, we propose a novel generalized end-to-end representation learning architecture, dubbed *Convolutional Dictionary Pair Learning Network* (CDPL-Net) in this paper, which integrates the learning schemes of the CNN and dictionary pair learning into a unified framework. Generally, the architecture of CDPL-Net includes two convolutional/pooling layers and two dictionary pair learning (DPL) layers in the representation learning module. Besides, it uses two fully-connected layers as the multi-layer perception layer in the nonlinear classification module. In particular, the DPL layer can jointly formulate the discriminative synthesis and analysis representations driven by minimizing the batch based reconstruction error over the flatted feature maps from the convolution/pooling layer. Moreover, DPL layer uses $l_1$-norm on the analysis dictionary so that sparse representation can be delivered, and the embedding process will also be robust to noise. To speed up the training process of DPL layer, the efficient stochastic gradient descent is used. Extensive simulations on real databases show that our CDPL-Net can deliver enhanced performance over other state-of-the-art methods.

## 1 INTRODUCTION

Image representation learning has been a core research topic in the areas of pattern recognition and computer vision, because learned salient features or representation directly influence the representation performance of learning systems. Several popular representation learning methods includes the sparse representation by dictionary learning (DL) [7-9][32-33], low-rank representation (LRR) [27-29] and the convolutional neural networks (CNN) based deep frameworks [24-25][31][34-35], etc. Notably, the CNN based deep models have been proved to be effective for image understanding, and have established the state-of-the-art performance on image recognition [11]. It is noteworthy that the CNN based deep models automatically extract effective deep representations from samples in an end-to-end manner, while both DL and LRR based methods are shallow single-layer learning models. In this paper, we aim to explore how to seamlessly integrate two learning systems into one unified framework for the discriminative representation learning.

Dictionary leaning (DL) is a long-standing representation learning technique, which has been successfully applied to a variety of image processing and recognition applications [4-10][26][37-44]. DL obtains the sparse representation of samples via a linear combination of atoms in a dictionary. Classical DL algorithms include *K-Singular Value Decomposition* (K-SVD) [4], *Discriminative K-SVD* (D-KSVD) [7], *Label-Consistent K-SVD* (LC-KSVD) [1], *Fisher Discrimination Dictionary Learning* (FDDL) [8], *Dictionary Learning with Structured Incoherence* (DLSI) [9], *Structured Analysis Discriminative Dictionary Learning* (ADDL) [12], *Projective Dictionary Pair Learning* (DPL) [10] and *Low-rank Shared Dictionary Learning* (LRSDL) [19], etc. Compared with the other existing methods, both DPL and ADDL extend the regular DL into the dictionary pair learning, i.e., learning a synthesis dictionary and an analysis dictionary jointly to analytically code data.

Due to the increasing attention of deep learning [34] and limited representation learning ability of single-layer dictionary leaning on the complex datasets, researchers have transformed to explore the settings of combining CNN and DL or proposing the deep learning based DL models directly, e.g., [2-3][32]. The recently proposed deep models usually have multiple layers, and recent works have shown that deeper architectures can be built from dictionary leaning [13]. For example, Chun *et al.* have presented a block proximal Gradient method by a majority for the convolutional dictionary leaning (CDL) [13]. Hu *et al.* developed a nonlinear DL method for image classification [14]. Tang et al. proposed a deep Micro-dictionary learning and coding network (DDLCN) for image representation and classification [21], which simply includes the feature extraction layer and multiple DL layers together. Singhal *et al.* have also proposed a deep dictionary leaning network DDL for image classification [22], which performs multiple DL for feature coding, followed by reconstruction error minimization. Wang *et al.* have proposed an object detection system by unifying the dictionary pair learning with convolutional feature learning [3], where the learned dictionary is used as a classifier yet only one layer.

It is worth noting that although certain deeper features or representations can be obtained by aforementioned deep DL methods, they still suffer from some drawbacks. First, they are typically not end-to-end deep DL frameworks. Prior to performing the deep DL, feature extraction operations are usually required to extract valid features from original data. In other words, the qualities of extracted features are greatly subjective to the learning ability and recognition accuracy. Second, they only simply add together multiple shallow DL layers, while the classification information cannot be extracted from data in multiple dimensions. To address this issue, existing methods have used a deep network model to extract features and then use the dictionary learning as a classifier to combine classification, but how to seamlessly integrate the deep network with DL for joint classification needs further investigation. In this way, it remains very challenging to seamlessly integrate the deep network and DL into a joint representation learning and classification framework. Third, existing DL based deep models are greatly confined by the costly model training methods, because each iteration needs to perform optimization over all training samples.

---
[1] School of Computer Science and Technology, Soochow University, China, emails: daitusun@gmail.com, cszzhang@gmail.com
[2] Key Laboratory of Knowledge Engineering with Big Data (Ministry of Education) & School of Computer Science and Information Engineering, Hefei University of Technology, Hefei, China, emails: yeungwangresearch @gmail.com, eric.mengwang@gmail.com
[3] School of Information Science and Technology, University of Science and Technology of China, Hefei, China, email: zhazj@ustc.edu.cn
[4] Department of Electrical and Computer Engineering, National University of Singapore, Singapore, email: eleyans@nus.edu.sg

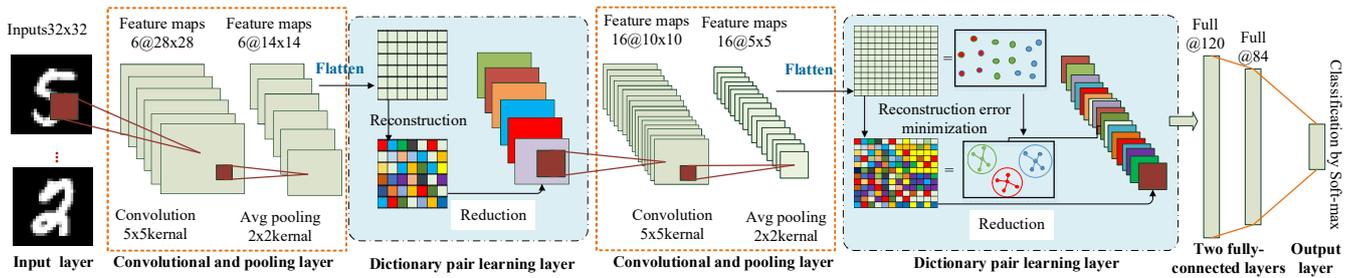

**Figure 1:** The architecture of our proposed CDPL-Net for image recognition.

Inspired by the complementary characteristics of CNN and dictionary pair learning on the learning mechanisms, this paper mainly investigates how to seamlessly integrate them together to improve the image representation and recognition abilities. The main contributions of this work are summarized as follows:

1. Technically, we propose a new and effective end-to-end deep dictionary learning framework, termed *Convolutional Dictionary Pair Learning Network* (CDPL-Net), which can make discriminative deep dictionary pair learning and effective image classification simultaneously. Specifically, the framework of CDPL-Net is constructed by two convolutional/pooling layers and two dictionary pair learning (DPL) layers in the representation learning module. Subsequently, two fully-connected layers are used for the nonlinear classification. The whole learning architecture are trained in an end-to-end manner.

2. To improve the representation learning ability in the DPL layer, CDPL-Net uses the $l_1$-norm on the analysis dictionary so that sparse representation can be delivered, and this can also make the embedding process robust to noise in data. To facilitate the optimization and avoid using the complementary data matrix to save time, we propose to compute a common synthesis dictionary and a common analysis dictionary over various classes.

3. Different from existing so-called deep DL techniques, CDPL-NET integrates the DPL process into convolutional neural network as a new layer for image representation by reconstruction. That is, sparse representation can be trained after pooling layer. Based on minimizing the reconstruction error within each batch and learning an overall synthesis dictionary for all classes, the mini-batch stochastic gradient descent method is used to speed up the training process of our network. In addition, CDPL-Net can extract features with multiple dimensions by the convolutional and polling layers, and use the involved DPL layers to integrate feature information of various dimensions to obtain an end-to-end convolutional dictionary pair learning network. Moreover, the classification process of CDPL-Net does not depend on the feature of original data when classifying data.

The paper is outlined as follows. Section 2 reviews the related works. Section 3 introduces the proposed CDPL-Net framework. In Section 4, we describe the experimental results and analysis. Finally, the paper is concluded in Section 5.

## 2 RELATED WORK

In this section, we briefly review the closely related algorithms that are closely related our proposed CDPL-Net framework.

### 2.1 Review of LeNet-5

We first review LeNet-5 briefly. The LeNet-5 architecture consists of two sets of convolutional and average pooling layers, followed by a flattening convolutional layer, then two fully-connected layers and finally a soft-max classifier. The input is a 32×32 grayscale image which passes through the first convolutional layer with 6 feature maps or filters having size 5×5 and a stride of one. Each unit in each feature is connected to a 5×5 neighborhood in the input, and the size of feature maps is 28×28. Then LeNet-5 uses the average pooling layer or sub-sampling layer with a filter size 2×2 and a stride of 2, and the resulting dimensions of feature maps are 14×14×6. The second convolutional layer has 16 feature maps having size 5×5 and a stride of 1. In this layer, only 10 out of 16 feature maps are connected to 6 feature maps of previous layer. The fourth layer is also an average pooling layer with filter size 2×2 and a stride of 2. This layer is the same as the second layer (S2) except it includes 16 feature maps so the output will be reduced to 5×5×16. The fifth layer represents a fully-connected convolutional layer with 120 feature maps each of size 1×1. Each of the 120 units in this layer is connected to all the 400 nodes (5×5×16) in previous layer. The sixth layer is also a fully-connected layer with 84 units. Finally, there is a fully-connected soft-max output layer. The structure of LeNet-5 is illustrated in Figure 2.

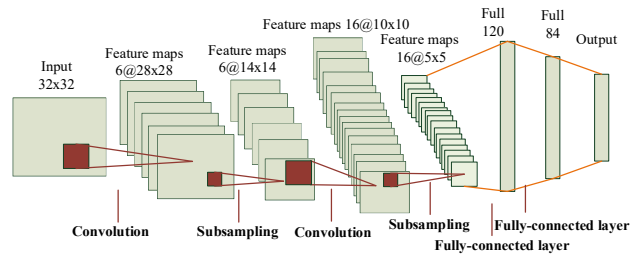

**Figure 2.** Architecture of LeNet-5 framework for character recognition.

### 2.2 Dictionary Learning and Dictionary Pair Learning

Let $X = [x_1, \cdots x_i, \cdots x_N] \in \mathbb{R}^{m \times N}$ denote a collection of training samples belonging to $c$ classes, where $x_i$ is the $i$-th sample in $X$, $m$ is the original dimensionality and $N$ is the number of samples. Then, DL computes a reconstructive dictionary $D = [d_1, \cdots d_K] \in \mathbb{R}^{m \times K}$ of $K$ atoms to obtain the sparsest representation $S = [s_1, \cdots s_N] \in \mathbb{R}^{K \times N}$ based on $X$ from the following optimization problem:

$$\langle D, S \rangle = \arg\min_{D,S} \|X - DS\|_F^2 + \lambda \|S\|_p, \quad (1)$$

where $\|X - DS\|_F^2$ is the reconstruction error, and $\lambda > 0$ is a positive scalar constant. $\|S\|_p$ is the $l_p$-norm regularization, where $p = 0/1$ corresponds to the widely-used $l_0$-norm/$l_1$-norm.

Different from the traditional settings of DL, DPL is modeled as a structured learning formulation aiming at calculating a synthesis discriminative dictionary $D = [D_1, \cdots D_i, \cdots D_c] \in \mathbb{R}^{n \times K}$ and an analysis discriminative dictionary $P = [P_1; \cdots P_i; \cdots P_c] \in \mathbb{R}^{K \times n}$ jointly through dictionary pair learning. Let $X = [X_1, X_2, \cdots, X_c] \in \mathbb{R}^{m \times N}$ denote the training sample sets from $c$ classes, where $X_l$ is the sample set of class $l$, including $N_l$ training samples, i.e., $N=\sum_{l=1}^{c}N_l$, then the objective function of DPL is defined as follows:

$$\langle P^*, D^* \rangle = \arg\min_{P,D} \sum_{l=1}^{c} \left\| X_l - D_l P_l X_l \right\|_F^2 + \lambda \left\| P_l \overline{X_l} \right\|_F^2, \ s.t. \|d_i\|_2^2 \leq 1, \quad (2)$$

where $\overline{X_l}$ is the complementary data matrix of $X_l$ in $X$ excluding $X_l$ itself from $X$. $D_l = [d_1, \cdots d_k] \in \mathbb{R}^{n \times K_l}$ and $P_l \in \mathbb{R}^{K_l \times n}$ denote the synthesis sub-dictionary and analysis sub-dictionary trained based on the class $l$, respectively. $K_l$ is the number of atoms according to class $l$. The constraint $\|d_i\|_2^2 \leq 1$ can avoid the trivial solution $P_l = 0$ and make the problem stable. DPL is a promising single-layer data representation and classification framework, however it cannot be directly integrated into the deep learning framework as a dictionary pair learning layer, since it learns the synthesis sub-dictionaries for each class Thus, DPL cannot be optimized by mini-batch stochastic gradient descent algorithm. Besides, DPL cannot deliver the sparse representation due to the fact that it does not impose a sparse constraint on the codes. In this paper, we mainly explore how to integrate the dictionary pair learning into the CNN based deep models.

## 3 CONVOLUTIONAL DICTIONARY PAIR LEARNING NETWORK (CDPL-Net)

We introduce the architecture and modules of our proposed convolutional dictionary pair learning network (CDPL-Net) for effective image representation and recognition.

### 3.1 The Whole Framework

We illustrate the general architecture of CDPL-Net in Fig.1, from which it is clear that the architecture of CDPL-Net is made of one input layer, multiple hidden layers and one output layer. The input includes the original images, and the dimensionality of input images is $h_{in} \times w_{in} \times c_{in}$, where $h_{in}$ and $w_{in}$ are their height and width, and $c_{in}$ is the number of channels in the images. If $c_{in} = 1$, the input is grayscale image, and $c_{in} = 3$ for RGB image. To train a model, the original images of a batch are feed into the network, so the dimension of the input $X_{in}$ is $b \times c_{in} \times h_{in} \times w_{in}$, where $b$ is batch size.

In this paper, the inputted image size is $32 \times 32$ with 1 channel, i.e., grayscale. Besides, since we focus on presenting our basic idea rather than designing a complex deep model, a simple deep neural network model is constructed to capture the hidden representations by integrating two projective dictionary pair learning (DPL) layers into the architecture of LeNet-5. As such, our CDPL-Net contains 10 layers for image representation, including one input layer, two convolutional layers, two pooling layers, two dictionary pair learning layers, two fully-connected layers and one output layer.

We describe the parameters in each layer of CDPL-Net in Table I. The first convolutional layer C1 has 6 feature maps, uses $5 \times 5$ kernel for each feature map, has 6 different bias and 6 different $5 \times 5$ kernels, which includes 156 trainable parameters. The first pooling layer P2 includes six feature maps and uses a $2 \times 2$ kernel for each feature map. The first DPL layer DPL3 has $K$ atoms. By the DPL layer, the output of P2 can be sparsely represented and then inputted to subsequent convolution layer after the activation function. The second convolution layer C4 has 16 feature maps, uses $5 \times 5$ kernel for each feature map, has 16 features and the neuron of each output feature map connects with some areas of $5 \times 5$ pixels at layer DPL3. The second pooling layer P5 has 16 feature maps and us-es $2 \times 2$ kernel for each feature map. The second DPL layer DPL6 has $K$ atoms of outputted feature maps that are sparsely represented in the layer P5. The first fully-connected layer F8 selects 120 neurons, and the second fully-connected layer F9 selects 84 neurons. Finally, the output layer has $c$ neurons for $c$ classes.

**Remarks**: The main difference between our CDPL-Net and LeNet-5 is that the convolutional and polling layers are connected to the DPL layers seamlessly. Compared with existing DL models, our CDPL-Net do not need to carefully select the features used, the feature extraction function of deep network can be used, meanwhile strong image representation and reconstruction abilities can be enhanced by the convolutional dictionary pair learning.

In this paper, we consider the classification problem of over $c$-classes. For our CDPL-Net, it considers using the cross entropy as the loss function, which is defined as

$$L = \sum_{j=1}^{N} \sum_{i=1}^{c} -p_{ji} \log(\widehat{p_{ji}}), \ where \ \hat{p} = \text{softmax}(f), \quad (3)$$

where $f$ is the outputted tensor data from the last fully-connected layer, $p_{ji}$ denotes the probability that the sample $j$ belongs to the category $i$, $\widehat{p_{ji}}$ denotes the predicted probability that sample $j$ is classified as the category $i$. In what follows, we will describe the convolutional/pooling layer and the DPL layer in detail.

### 3.2 Convolutional and Pooling Layer

In the input layer, the original dimensionality of the inputted image data $X_{in}$ is $b \times c_{in} \times h_{in} \times w_{in}$. We first extract the region-based deep features by feeding given image data into the convolutional layer. The extracted deep representations are denoted as $W_c \otimes X_{in}$, where $\otimes$ is a set of operations of convolution, average polling and ReLU activation function, and $W_c$ are the overall parameters. Based on the convolutional and polling layers for feature extraction within each batch, we obtain an output as follows:

$$X_{out} = g(W_c \otimes X_{in}), \quad (4)$$

with dimension $b \times c_{out} \times h_{out} \times w_{out}$ for subsequent DPL layer, where $h_{out}$ and $w_{out}$ are the height and width of output feature map, $c_{out}$ is the number of channels of feature map and function $g(\bullet)$ can be represented by a convolutional operation and a polling operation.

**Table 1.** Layer Description of our CDPL-Net Framework

| Layers | Descriptions |
|---|---|
| Layer 1 Input layer | Number of feature map:1, Number of parameters:0<br>Number of trainable parameters:0 |
| Layer 2 [C1] Convolution | Number of feature map:6, Number of parameters:156<br>Number of trainable parameters:156 |
| Layer 3 [P2] Pooling | Number of feature map:6, Number of parameters:12<br>Number of trainable parameters:12 |
| Layer 4 [DPL3] | Dictionary pair learning<br>Number of feature map:6, Number of dictionary: $K$ |
| Layer 5 [C4] Convolution | Number of feature map:16, Number of parameters:1516<br>Number of trainable parameters:1516 |
| Layer 6 [P5] Pooling | Number of feature map:16, Number of parameters:32<br>Number of trainable parameters:32 |
| Layer 7 [DPL6] | Dictionary pair learning<br>Number of feature map:16, Number of dictionary: $K$ |
| Layer 8 [F8] fully-connected 1 | Number of feature map:120, Number of parameters: 48120, Number of trainable parameters:48120 |
| Layer 9 [F9] fully-connected 2 | Number of feature map:84, Number of parameters:1210<br>Number of trainable parameters:1210 |

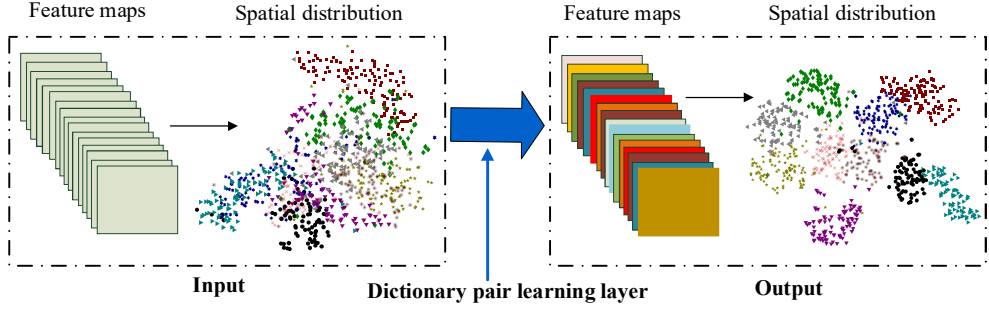

Figure 3. The schematic diagram of the spatial distribution of feature maps before and after the DPL layer on the MNIST database.

### 3.3 Projective Dictionary Pair Learning Layer

In the framework of CDPL-Net, the DPL layer follows the convolutional and polling layer. Assume that after the convolutional/pooling layer, the dimension of output $X_{out}$ is $b \times c_{out} \times h_{out} \times w_{out}$, where $b$ denotes the batch size, $c_{out}$ is the number of feature maps, $h_{out}$ and $w_{out}$ denote the height and width of each feature map, respectively. To integrate the dictionary pair learning process into LeNet-5 as the integrated layer, two improvements are considered. First, prior to learning the representations, the input of the dictionary pair learning layer needs to be converted from $[b, c_{out}, h_{out}, w_{out}]$ to $[h_{out} \times w_{out}, b \times c_{out}]$, where $h_{out} \times w_{out} = m$ and $b \times c_{out} = n$, so that each feature map can be transformed reversely as a sample for the dictionary pair learning. Second, we compute an overall projective analysis dictionary $P \in \mathbb{R}^{k \times n}$ for all the classes rather than individual analysis sub-dictionaries over each class. In addition, we also consider the sparse representation by using the $l_1$-norm. As such, the cost function for the dictionary pair learning layer is defined as follows:

$$\arg\min_{D,P} \sum_{l} \|X_l - DPX_l\|_F^2 + \beta \|P\|_1, \quad s.t. \quad \forall_i \|d_i\|_2^2 \leq 1, \quad (5)$$

where $X_l \in \mathbb{R}^{m \times n}$ is the output of each batch $l$ in previous convolutional and pooling payer and $P \in \mathbb{R}^{k \times n}$ denotes an overall projective analysis dictionary that can analytically code $X_l$. By minimizing the reconstruction error within each batch and learning an overall synthesis dictionary $P$, the mini-batch stochastic gradient descent method can be applied to solve the above problem, which can speed up the training process of our network. $\beta > 0$ is scalar constant. The $l_1$-norm constraint on the synthesis dictionary $P$, i.e., $\|P\|_1$, can ensure the sparsity of the resulted representation.

After DPL layer, the reconstruction of $X_i$ is obtained as $DPX_i$. To link with the other layers behind the DPL layer, the dimension of $DPX_i$ needs to be changed from $[m, n]$ to $[b, c_{out}, h_{out}, w_{out}]$ conversely. By following [23], we regard the DPL layer with an activate function as the scale exponential linear units (SELUs). It is noteworthy that the SELU activation function is defined as

$$selu(x) = \lambda \begin{cases} x & if \ x > 0 \\ \alpha e^x - \alpha & if \ x \leq 0 \end{cases}, \quad (6)$$

where $\alpha = 1.6732632$ and $\lambda = 1.0507009$ [23]. To show the learning ability of the DPL layer, we use a schematic diagram to show the feature maps before and after the DPL layer in Figure 3.

## 4 EXPERIMENTS

We mainly evaluate our CDPL-Net for image representation and classification. The performance of CDPL-Net is mainly compared with several traditional DL methods including the sparse representation based classification (SRC) [5], DLSI [9], D-KSVD [7], LC-KSVD [1], FDDL [8], DPL [10], LRSDL [19] and ADDL [12], and four related deep learning models, including deep sparse coding network (SCN) [6], DDL [22], and DDLCN [21]. For image representation and classification on each database, we split it into a training set and a test set. We perform all simulations on a PC with Intel (R) Core (TM) i7-7700 CPU @ 3.6 GHz 8G.

Table 1. Descriptions of Used Real-world Image Datasets

| Dataset Name | # Samples | # Dim | # Classes |
|---|---|---|---|
| AR face [17] | 2600 | 1024 | 100 |
| CMU PIE face [18] | 11554 | 1024 | 68 |
| MNIST [16] | 70000 | 784 | 10 |
| Fashion-MNIST [15] | 70000 | 784 | 10 |

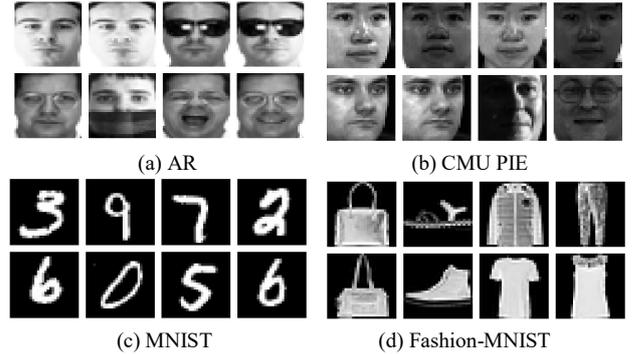

(a) AR  (b) CMU PIE

(c) MNIST  (d) Fashion-MNIST

Figure 4. Sample images of evaluated real-world image databases.

### 4.1 Descriptions of used Datasets

We evaluate the performance of each learning method on four popular and standard datasets. Details of the evaluated image datasets are shown in Table II. AR face database has over 4000 color images of 126 people. Each person has 26 face images taken during two sessions. Following the common procedures, the face set including images of 50 males and 50 females is applied. CMU PIE face dataset [18] contains 68 persons with 41368 face image as a whole. Following [12], 170 near frontal images per person are used. This face set has five near frontal pose (C05, C07, C09, C27, and C29) and all face images have different illuminations, lighting and expression. As is common practice, all the images of AR and CMU PIE are resized into 32×32 pixels in grayscale format. MNIST is a handwritten digit dataset [16] that has a training set of 60,000 examples and a test set of 10,000 examples, where the samples are 28×28 grayscale images. Fashion-MNIST [15] is a new dataset comprising of 28×28 grayscale images of 70,000 fashion products from 10 categories, with 7000 images per category. The training set has 60,000 images and the test set has 10,000 images. Fashion-MNIST is a more difficult task than MNIST for evaluating the machine learning algorithms. Some sample images of the evaluated databases are illustrated in Figure 4.

### 4.2 Convergence Analysis

We would like to provide some numerical results to analyze the convergence behavior of our proposed CDPL-Net. We analyze the convergence behavior by describing the loss function values. AR, CMU PIE, MNIST and Fashion-MNIST datasets are used. For AR, we will iterate 200 epochs for training, select half of the images for training randomly and use the rest for testing. For CMU PIE, we iterate 100 epochs for training, randomly select half of the images for training and test on the rest. For MNIST and Fashion-MNIST, we will iterate 30 epochs for training. The results are presented in Figure 5. We find that the loss function value of our CDPL-Net decreases rapidly. For smaller datasets such as AR and CMU PIE, the loss function value jumps out of the local optimal solution to look for a smaller solution of the loss function. For larger datasets such as MNIST and Fashion-MNIST, the loss is non-increasing in iterations and finally converges.

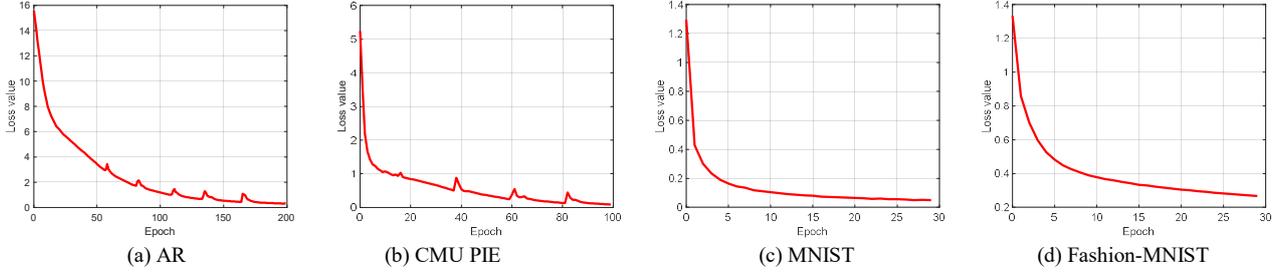

(a) AR  (b) CMU PIE  (c) MNIST  (d) Fashion-MNIST

**Figure 5.** Convergence behavior of our CDPL-Net, where the *x*-axis is the number of epochs and the *y*-axis represents the loss function values.

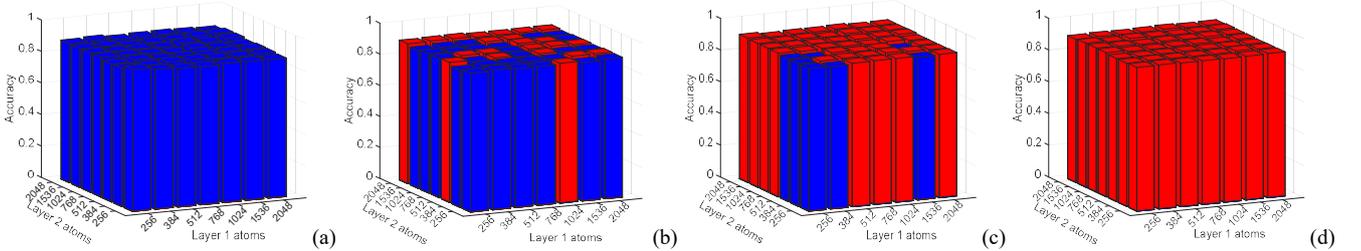

(a)  (b)  (c)  (d)

**Figure 6.** Parameter sensitivity of CDPL-Net on Fashion-MNIST database, where (a) the effects of tuning dictionary number on the performance by fixing $\beta=1\times10^{-2}$ ; (b) the effects of tuning dictionary number on the performance by fixing $\beta=1\times10^{-3}$ ; (c) the effects of tuning dictionary number on the performance by fixing $\beta=1\times10^{-4}$ , (d) the effects of tuning dictionary number on the performance by fixing $\beta=1\times10^{-5}$ .

**Table 3.** Comparison of Recognition Results on CMUPIE

| Methods | 50% train | 75% train |
|---|---|---|
| SRC [5] | 90.30% | 91.58% |
| DKSVD [7] | 92.31% | 92.92% |
| LC-KSVD1 [1] | 93.67% | 95.01% |
| LC-KSVD2 [1] | 94.49% | 95.91% |
| DLSI [9] | 95.48% | 96.13% |
| FDDL [8] | 95.38% | 96.00% |
| DPL [10] | 94.20% | 95.00% |
| LRSDL [19] | 96.26% | 96.85% |
| ADDL [12] | 95.90% | 96.30% |
| SCN-2 [6] | 96.37% | 97.53% |
| CDPL-Net (no DPL layers) | 96.20% | 97.44% |
| **Our CDPL-Net** | **97.70%** | **98.48%** |

### 4.3 Parameter Selection Analysis

We analyze the parameter sensitivity of CDPL-Net in this study. Since the parameter selection issue is still an open issue, we use a heuristic way to select the most important parameters. CDPL-Net has two DPL layers and one tunable parameter $\beta$ . For each pair of parameters, the classification results are averaged based on 10 splits of training/test samples with the varied parameter $\beta$ from $\{1\times10^{-5}, 1\times10^{-4}, 1\times10^{-3}, 1\times10^{-2}\}$ and number of dictionary atoms from $\{256, 384, 512, 768, 1024, 1536, 2048\}$. In this study, Fashion-MNIST dataset is applied as an example. The parameter selection results are illustrated in Figure 6, where four groups of results are illustrated. More specifically, the accuracy rates lower than 90% are highlighted using blue color, while those higher than 90% are denoted by red color. As can be seen, CDPL-Net works well in a wide range of the parameter selections in each group, which means that our proposed CDPL-Net is insensitive to the parameters, and a small $\beta$ tends to deliver higher recognition accuracy.

### 4.4 Face Recognition on CMU PIE

We compare the performance of CDPL-Net with those of SRC, LC-KSVD, DLSI, DPL, LRSDL, ADDL and SCN. To demonstrate the positive effects of incorporating the DPL layer into the deep models, we also add the results of our CDPL-Net without using the two DPL layers, which is called CDPL-Net (no DPL layers). We normalize the pixel value of the images into 0~1. For face recognition, we randomly select 50% and 75% of the images for model training and the remaining images for testing. The averaged results over 10 random splits are reported for evaluations. For traditional DL methods, the dictionary contains 2720 atoms, corresponding to an average of 40 items each category. For deep dictionary learning method SCN, we use the SCN-2 model with 100 epochs [6]. For our CDPL-Net, we use the cross entropy as loss function, applying the Adam algorithm [20] for training, where the learning rate is 0.0003 and each batch has 64 samples. The dictionaries in the two

DPL layers of CDPL-Net have 512 and 1024 atoms, respectively. We will iterate 30 epochs for training. For CDPL-Net (no DPL layers), the other setting are exactly the same as CDPL-Net and is used in CDPL-Net. The results of compared methods are given in Table 3. We can find that CDPL-Net is superior to its competing methods for face recognition in most cases. One can also find that our proposed CDPL-Net with the two DPL layers can deliver a 1.5% improvement over the setting of CDPL-Net without DPL layers, which implies that integrating the DPL layers into existing deep models can indeed improve the performance.

### 4.5 Handwriting Digit Recognition on MNIST

In this study, we evaluate each method for recognizing the handwritten digits of the MNIST database. In this study, the pixel values of digital images are normalized into 0~1. For traditional representation learning methods, i.e., SRC, DLSI, D-KSVD, LC-KSVD, FDDL, DPL, LRSDL and ADDL, the dictionary contains 2000 atoms, corresponding to an average of 200 items of each class, and we average the results over 10 random splits of training/test samples. For deep DL method SCN, we use the SCN-4 model with 30 epochs [6]. For CDPL-Net and CDPL-Net (no DPL layers), we use the cross entropy as loss function, use the Adam algorithm for training, where the learning rate is 0.0003 and each batch has 64 samples. The number of atoms in DPL layers of CDPL-Net is set to 1024. We iterate 30 epochs for training and $\beta = 0.0001$ is set in our CDPL-Net. The results are shown in Table 4. We can observe that our CDPL-Net outperforms all its competing methods in most cases, which demonstrates the effectiveness of our framework.

**Table 4.** Recognition Results on MNIST and Fashion-MNIST

| Methods | MNIST | Fashion-MNIST |
|---|---|---|
| SRC [5] | 84.61% | 79.86% |
| DKSVD [7] | 84.69% | 78.38% |
| LC-KSVD1 [1] | 84.72% | 78.50% |
| LC-KSVD2 [1] | 85.88% | 79.27% |
| DLSI [9] | 88.39% | 79.48% |
| FDDL [8] | 87.93% | 80.67% |
| DPL [10] | 90.08% | 83.50% |
| LRSDL [19] | 87.80% | 81.99% |
| ADDL [12] | 88.90% | 82.10% |
| DDL [22] | 98.33% | - |
| SCN-4 [6] | 97.98% | 88.73% |
| DDLCN (100-100) [21] | 98.55% | - |
| CDPL-Net (no DPL layers) | 98.64% | 87.46% |
| **Our CDPL-Net** | **98.98%** | **90.69%** |

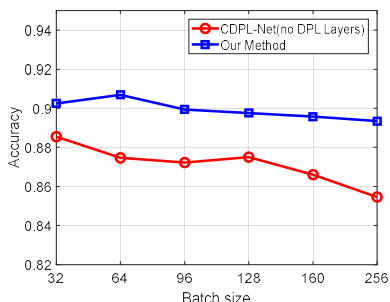

**Figure 7.** Accuracy on Fashion-MNIST with different batch sizes.

### 4.6 Products Recognition on Fashion-MNIST

Fashion-MNIST database shares the same image size, data format and the structure of training and testing splits as original MNIST dataset, but its task is more difficult than MNIST. In this study, the pixel value of fashion product images are also normalized into 0~1. For traditional representation learning methods, the atom number is set to 2000, corresponding to an average of 200 items of each fashion product class. For the deep DL method SCN, we also use the SCN-4 model with 30 epochs [6] for fair comparison. The loss function, training algorithm, learning rate, batch size and dictionary sizes in the DPL layers are set the same as those on MNIST. The averaged recognition results are shown in Table 4. From the results, we find that CDPL-Net outperforms its competing methods. Moreover, CDPL-Net with DPL layers obtains a 3% improvement over the CDPL-Net without DPL layers, which once again demonstrates the positive contributions of the DPL layers.

In addition, we visualize the impact of different batch sizes on the performance of CDPL-Net. We illustrate the comparison results of CDPL-Net and CDPL-Net (no DPL layers) by varying the batch size in Figure 7. It is clear that CDPL-Net performs better than CDPL-Net (no DPL layers) and is also more stable than CDPL-Net (no DPL layers). Thus, one can conclude that the integrating the DPL layers into the deep convolutional architecture is beneficial to improve the recognition performance, and enhance the stability of the deep representation learning algorithm.

### 4.7 Data Visualization

**Visualization of features.** We first use the MNIST dataset to show the spatial distribution of convolutional features and reconstruction features in each layer of our CDPL-Net. After training CDPL-Net, only 1000 images are selected for clear observation. For data visualization, the output of the convolutional and pooling layer are called convolutional features, while the features after the DPL layers are called reconstruction features. After the training, we can obtain two convolutional features and two reconstruction features. For the four features, the *t-Distributed Stochastic Neighbor Embedding* (t-SNE) [30] algorithm is applied to reduce the dimension of features to 3 and visualize them in a 3D space. The convolutional features and reconstruction features are visualized in Figure 8. We see that the reconstruction features are much better than convolutional features in terms of high inter-class separation and intra-class compactness. This also implies that the used DPL layers can effectively work on convolutional features and moreover it clearly encourages features to be discriminative in the resulting feature space, which will be beneficial to subsequent image recognition.

**Visualization of the confusion matrix.** In addition to presenting the above recognition results, we also show the confusion matrices on the test sets of Fashion-MNIST and MNIST in Figure 9. From the visualization results, we can find that the recognition accuracies on MNIST are usually higher than those on Fashion-MNIST, and the confusion occurs within the sixth class of MNIST and seventh class of Fashion-MNIST, respectively.

For MNIST and Fashion-MNIST datasets, we first use the training set to train CDPL-Net and then obtain the features of the test set at relevant layers. The number k of clusters on test set is set to the number of categories. For each setting, we average the numerical clustering accuracy (AC) values [36] over 20 random initializations for the k-means clustering algorithm. The number of atoms in the DPL layers of CDPL-Net is set to 1024. We iterate 30 epochs for training and $\beta = 10^{-4}$ is set for CDPL-Net. We report the averaged and highest AC values over different runs in Table 5. We find that the clustering results on the DPL layer features are superior to those on the convolutional/polling features obviously. This can also

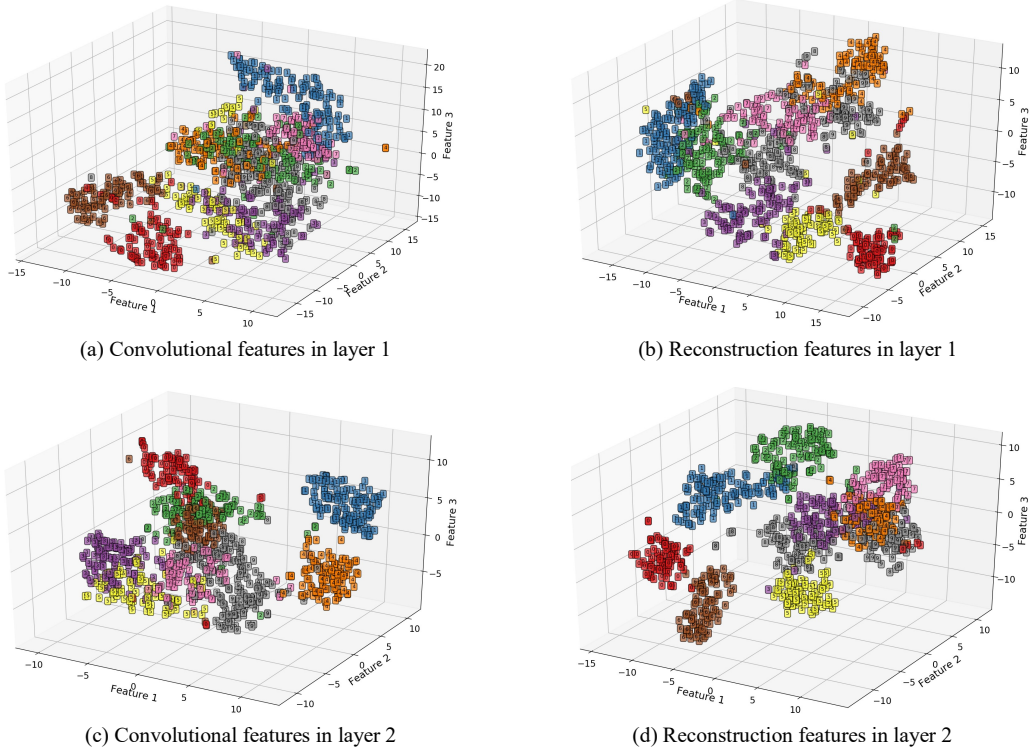

(a) Convolutional features in layer 1     (b) Reconstruction features in layer 1

(c) Convolutional features in layer 2     (d) Reconstruction features in layer 2

**Figrue 8.** Visualization of convolutional features and reconstruction features obtained in the two layers of CDPL-Net on the MNIST database.

(a) MNIST     (b) Fashion-MNIST

**Figure 9.** Confusion matrix on MNIST and Fashion-MNIST datasets.

**Table 5.** The clustering AC with different layer features on the MNIST and Fashion-MNIST datasets.

| Layers | MNIST | | Fashion-MNIST | |
|---|---|---|---|---|
| | Mean AC | Max AC | Mean AC | Max AC |
| Layer 3 [P2] | 56.15% | 60.80% | 50.88% | 56.73% |
| Layer 4 [DPL3] | 61.57% | 67.22% | 53.25% | 60.47% |
| Layer 6 [P5] | 77.68% | 86.46% | 60.88% | 64.89% |
| Layer 7 [DPL6] | **86.95%** | **93.79%** | **66.95%** | **75.20%** |

prove that integrating DPL layers can indeed obtain more separable subspaces than the convolutional/polling features.

## 5 CONCLUSION AND FUTURE WORK

In this paper, we proposed a new and effective convolutional dictionary pair learning network called CDPL-Net, which seamlessly integrates the convolutional networks and the projective dictionary pair learning into one unified deep dictionary learning framework.

In the proposed framework, the DPL layer could employ the flatted convolutional features as inputs for discriminative sparse representation learning within each batch. Meanwhile, the convolution and pooling layer can be performed based on the reconstructed data for calculating new feature maps. Based on interacting the DPL layers and convolutional/pooling layers, the representation learning ability is greatly improved. Extensive experiments and results demonstrated the effectiveness of our proposed network on several popular image databases. In this paper, we have mainly presented an approach to integrate the DPL unit into a simple deep convolutional network. However, we believe that the DPL unit could be potentially incorporated into any other deeper networks as an integrated layer with deeper structures, e.g., VGG [24], and further extended to other related fields, e.g., video surveillance modeling.

## ACKNOWLEDGEMENTS

This paper is partially supported by the National Natural Science Foundation of China (61672365, 61732008, 61725203, 61622305, 61871444 and 61806035) and the Fundamental Research Funds for Central Universities of China (JZ2019HGPA0102). Zhao Zhang is the corresponding author of this paper.